\pdfoutput=1
\documentclass[11pt]{article}

\usepackage[review]{EACL2023}

\usepackage{times}
\usepackage{latexsym}

\usepackage[T1]{fontenc}
\usepackage[utf8]{inputenc}

\usepackage{microtype}

\usepackage{inconsolata}

\usepackage{xcolor}
\usepackage{hyperref}
 \definecolor{darkblue}{rgb}{0, 0, 0.5}
  \hypersetup{colorlinks=true, citecolor=darkblue, linkcolor=darkblue, urlcolor=darkblue}

\usepackage{xstring}

\usepackage{graphicx}
\usepackage{tabularx}
\usepackage{soul}
\usepackage{algorithm}
\usepackage{algorithmic}
\usepackage{bm}
\usepackage{multirow}
\usepackage{color}
\usepackage{amsthm,amsmath,amssymb}
\usepackage{mathrsfs}
\usepackage{graphicx}
\usepackage{verbatim}
\usepackage{enumerate}
\usepackage{array}
\usepackage{xcolor,colortbl}
\makeatletter
\newcommand{\thickhline}{%
    \noalign {\ifnum 0=`}\fi \hrule height 1pt
    \futurelet \reserved@a \@xhline
}
\title{Is Crowdsourcing Breaking Your Bank? Cost-Effective Fine-Tuning of Pre-trained Language Models with Proximal Policy Optimization}

\author{Shuo Yang \\
  Technical University of Munich \\
  \texttt{shuo.yang@tum.de} \\\And
  Gjergji Kasneci \\
  Technical University of Munich \\
  \texttt{gjergji.kasneci@tum.de} \\}

\begin{document}
\maketitle
\begin{abstract}
    Wide usage of ChatGPT has highlighted the potential of reinforcement learning from human feedback. However, its training pipeline relies on manual ranking, a resource-intensive process. To reduce labor costs, we propose a self-supervised text ranking approach for applying Proximal-Policy-Optimization to fine-tune language models while eliminating the need for human annotators.
    Our method begins with probabilistic sampling to encourage a language model to generate diverse responses for each input. We then employ TextRank and ISODATA algorithms to rank and cluster these responses based on their semantics. Subsequently, we construct a reward model to learn the rank and optimize our generative policy.
    Our experimental results, conducted using two language models on three tasks, demonstrate that the models trained by our method considerably outperform baselines regarding BLEU, GLEU, and METEOR scores. Furthermore, our manual evaluation shows that our ranking results exhibit a remarkably high consistency with that of humans. This research significantly reduces training costs of proximal policy-guided models and demonstrates the potential for self-correction of language models.
\end{abstract}

\section{Introduction}
    With the advancement of natural language processing, contemporary pre-trained language models (PLMs) have demonstrated significant commercial value due to their widespread adoption in sectors such as education, healthcare, and finance~\citep{edunov-etal-2019-pre, Kaili-Survey-PLM-2022}. However, due to the shortcut learning~\citep{geirhos2020shortcut}, degeneration~\citep{Holtzman2020The}, and other complicated reasons, PLMs often generate topic-irrelevant or unhelpful information, resulting in a loss of resources and reliability~\citep{Weidinger2021EthicalAS}. As an existing optimization, models trained through reinforcement learning from human feedback (RLHF)~\citep{wang2022self} are continually supervised by human-ranked data during the training. Therefore, they demonstrate higher performance and reliability across diverse tasks such as dialogue, question-answering, and machine reading comprehension.

    \begin{comment}
        \begin{figure}[t]
        \centering
            \includegraphics[width=7cm]{figures/chat.png}
            \caption{An example sampled from the DailyDialogue data set. The GPT-2 fine-tuned by our methods provide a more natural and fluent response.}
            \label{fig:example}
    \end{figure}
    \end{comment}
    \begin{comment}
    Specifically, the RLHF pipeline mainly includes three steps:
    1) fine-tuning a PLM on a prompt set as a question-answering bot;
    2) generating multiple answers for each question and then ranking them via crowdsourcing; and
    3) utilizing these rankings to train a reward model that evaluates the quality of answers for further fine-tuning the PLM.
    \end{comment}
    
    Although RLHF has been widely proven to be effective in improving the quality of generative models~\citep{9131765}, there are three limitations of applying RLHF:
    1) training costs of large-scale PLMs,
    2) lack of high-quality prompts for varying user intents~\cite{bodonhelyi2024user} and
    3) labor costs associated with crowdsourcing.
    
    Both of the first two limitations have been addressed with diverse solutions. Regarding the first limitation, contemporary lightweight PLMs such as Stanford Alpaca and ChatGLM~\citep{alpaca,zeng2023glm-130b} have achieved performance comparable to traditional large-scale PLMs with a hundred billion parameter, substantially reducing training costs.
    For the second limitation, prompt generation methods, e.g., self-instruction~\citep{wang2022self}, have presented solutions for automatically building instruction sets.
    However, for the last point, there needs to be more focus on exploring the utilization of self-supervised learning as a viable alternative to annotations on crowdsourcing platforms to address the challenge of substantial manual costs.

\begin{figure*}[t]
        \centering
            \includegraphics[width=16cm]{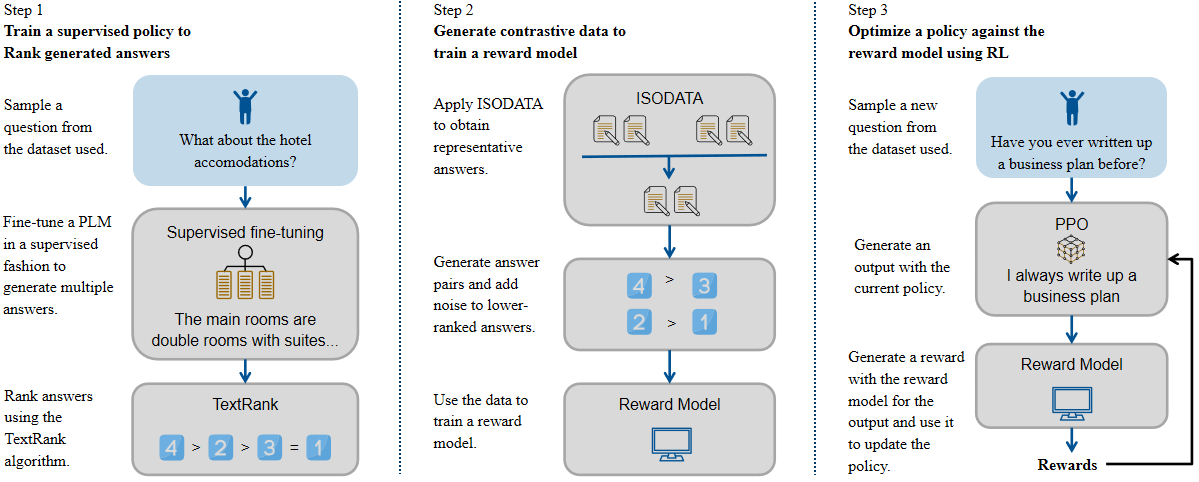}
            \caption{Our pipeline comprises three steps: 1) fine-tuning a language model to generate multiple candidate answers for a given question and using the TextRank algorithm to rank these answers; 2) filtering out non-representative answers using the ISODATA algorithm and training a reward model based on the remaining answers, and 3) scoring the ranked answers using the reward model and updating the generation policy via PPO. Note that we implemented the generative policy through a PLM in our study.}
            \label{fig:overview}
    \end{figure*}

    To achieve this, we propose a \underline{S}elf-supervised \underline{T}ext \underline{R}anking (STR) pipeline, simulating the generation of human-ranked data. We derive our theoretical and empirical foundations from two articles: \citet{chen2023teaching} demonstrated that language models can enhance generation quality through self-checking and correction. \citet{li-etal-2023-making} established the effectiveness of ensemble learning in assessing the rationality of various interpretations produced by a PLM when it employs different reasoning pathways to address the same question. Building upon these two contributions, we apply the proximal policy optimization (PPO) through the STR pipeline to enable language models to self-assess and self-supervise during fine-tuning~\citep{schulman2017ppo} as shown in Figure~\ref{fig:overview}, with the following three steps: 

    \begin{enumerate}
        \item \textbf{Ensemble learning-based text ranking.} 
        We follow the RLHF baseline, generating diverse answers for each question via a PLM. 
        After that, we apply a semantic similarity-based TextRank algorithm~\citep{mihalcea-tarau-2004-textrank, zhang2021unsupervised} to rank generated answers, distinguishing our work from previous efforts. We root our motivation in the theoretical assumption that if a PLM generates different answers to a given question, the semantics among reasonable answers should exhibit a stronger clustering tendency than irrational ones. This is because incorrect or unhelpful statements hallucinated by PLMs always involve various unrelated topics~\citep{zhang2023language}.
        \item \textbf{Extraction of representative answers.} 
        We then cluster answers via the Iterative Self-Organizing Data Analysis Technique Algorithm (ISODATA) and extract cluster centers to build answer pairs. The advantage of clustering is the reduced computational overhead and the avoidance of comparing semantically similar sentences. These constructed answer pairs will be utilized to train a reward model for assessing the quality of an answer. 
        \item \textbf{To update the generation policy.} We finally learn a reward model from the answer pairs. The reward model is then used to update our generative policy, which generates answers for evaluation.
    \end{enumerate}
    
    Our contributions are as follows:
    \begin{itemize}
        \item We propose a novel self-supervised text ranking method for simulating manual ranking in RLHF while eliminating human labor costs in fine-tuning PLMs.
        \item Our experimental results demonstrate that the proposed method significantly outperforms other fine-tuning approaches for two PLMs on three datasets.
        \item Our manual evaluation experiments demonstrate that our approach can considerably substitute human annotators for generating training data for future PPO-guided PLMs.
    \end{itemize}

\section{Related Work}
    \subsection{Pre-trained language models}
        Transformer-based PLMs~\citep{10.5555/3295222.3295349,     lewis-etal-2020-bart} have been widely used in assorted downstream tasks due to their versatility and excellent semantic extraction capabilities. Among them, ERNIE~\citep{zhang-etal-2019-ernie} incorporates knowledge from knowledge graphs to improve representation learning for NLU tasks. T5~\citep{ni-etal-2022-sentence} achieved transforming various NLP tasks into text-to-text transfer problems. 
        This paper uses the GPT-2 and GPT-Neo~\citep{radford2019language, gpt-neo} due to the extensive data sources they used in pre-training and their representative architectures stacked by attention layers. 

        Before inference, researchers often conduct fine-tuning by retraining particular parameters of a PLM on downstream datasets to adapt and optimize it for specific tasks~\cite{pfeiffer-etal-2020-adapterhub}.
        A fundamental fine-tuning approach involves training all the parameters with smaller learning rates. However, \citet{lin-etal-2020-exploring} demonstrated that by adding and fine-tuning additional 2-3\% parameters, PLMs could maintain a similar performance of the full fine-tuning.
        Furthermore, \citet{ben-zaken-etal-2022-bitfit} proposed a sparse-fine-tuning method where only the bias terms are being modified. \citep{hu2022lora} injected rank decomposition matrices into the Transformer architecture, reducing the trainable parameters for downstream tasks. However, these methods focus on optimizing model architecture or the size of trainable parameters while neglecting the effect of fine-tuning in enhancing the model's ability to self-correction.
        
        \subsection{Policy gradient}
        As a branch of reinforcement learning (RL)~\citep{minsky1961steps}, policy gradient algorithms~\citep{sutton1999policy} have been widely applied to NLP tasks, such as addressing the issue of gradient unavailability~\citep{yu2017seqgan} in Generative Adversarial Networks~\citep{goodfellow2020generative}. 
        As an off-policy improvement, \citet{schulman2017ppo} propose PPO, which computes the similarity between the generative and sampling policies as part of the objective function during training. 
        
        In the InstruchGPT~\cite{ouyang2022training}, a proxy model with the same architecture as the training PLM is employed to explore generation strategies for answers. However, updating the policy requires human feedback, significantly increasing the training cost. In this paper, we follow this work and present a cost-efficient solution.

        \subsection{Text ranking}
        Contemporary unsupervised text ranking methods rely on statistics, such as BLEU~\citep{papineni-etal-2002-bleu}, ROUGE~\citep{lin-2004-rouge}, METEOR~\citep{banerjee-lavie-2005-meteor},  and BERTScore~\citep{bert-score}, compute overlaps of n-grams or semantic information to rank answers based on their similarity to questions. However, while answers similar to the question are topic-relevant, their helpfulness to humans cannot be guaranteed.

        %Text ranking methods based on active learning. 
        %X employed workers to modify generated text and calculate edit distance, with smaller edit distance indicating higher text quality. 
        %Y compares outputs from multiple models for the same input and asks annotators to select the best one, evaluating these models' capability. 
        %However, these methods still incur manual annotation costs and are subject to subjective judgment in the annotation process.
    
\section{Methodology}
\subsection{Problem formulation}
    We aim to automatically rank text by leveraging the semantic information widely learned by PLMs during pre-training, thus implementing the PPO-guided training under unsupervised conditions to further fine-tune PLMs. Our study considers various natural language processing tasks as question-answering.

    Formally, given a parallel dataset consisting of a prompt $P$, a set of questions $Q=(q_1, ...,q_m)$, and their corresponding answers $A=(a_1, ...,a_m)$, we are going to fine-tune a question-answering PLM $\text{LM}_\phi$ and enhance its performance via PPO achieved by using ranked text. 
    
\subsection{How to generate diverse answers?}
    We initially follow the training paradigm of RLHF, wherein we fine-tune a PLM to produce distinct answers for a given question~\citep{ouyang2022training}. Specifically, to generate tokens for answering, we compute logits for each input token sequence $w_1, ..., w_{(i-1)}$ using $\text{LM}_\phi$. The probability of generating the $i$'th token $w_i$ as $w\in V$ is then given by $P_{\phi}(w|w_{1:i})$, where $V$ represents the vocabulary.  Subsequently, we apply a temperature function~\citep{ACKLEY1985147} with a small $\tau$, to shape the probability distribution:
    \begin{equation}
        P_{\phi}'(w_i)=\frac{P_{\phi}(w|w_{1:i})^{1/\tau}}{\sum_{w' \in V}P_{\phi}(w'|w_{1:i})^{1/\tau}}.
    \end{equation}
    To encourage the $\text{LM}_\phi$ to generate answers via diverse expression while mitigating potential text degeneration, we apply a top-p sampling with a high $p$ to conduct a subset of $V$ and sample token $w_i$ according to probability distribution $P_\phi''$:
    \begin{equation}
        P_{\phi}''(w_i) = \begin{cases}
        P_{\phi}'(w_i)/\sum_{w' \in V_p}P'_{\phi}(w') &\text{if }w_i \in V_p\\
        0 & \text{otherwise},
        \end{cases}
    \end{equation}
    where top-p vocabulary $V_p$ is the smallest set such that: $\sum_{w_i \in v_p} P(w_i|w_{1:i-1}) \geq p$.

    By repeatedly sampling from $P_{\phi}''(w_i)$, we generate multiple answers for each question in the dataset.
    
\subsection{How to rank generated answers?}
    \label{sec:rank}
    For a specific question, we assume that the semantic similarity among different answers can reflect the correctness of these answers in theory. Specifically, high-quality answers should exhibit more similarity in the hidden space than others. For example, given a question ``\emph{What does `apple' mean?}", possible answers include ``\emph{a kind of red fruit}", "\emph{a fruit rich in vitamins}", ``\emph{a US-based business}", and ``\emph{a company involved in electronic production}". Although there may not be a unique correct answer, in this case, the first two answers relate to ``apples", while the last two refer to ``Apple Inc.". Therefore, in theory, we could recognize their corresponding clusters in semantic space. Conversely, incorrect or irrelevant answers, such as ``\emph{a toxic substance.}" or ``\emph{I am full.}", would be more dispersed in the semantic space as the hallucinations~\citep{alkaissi2023artificial} always involve various unrelated topics or concepts~\citep{azamfirei2023large}. Therefore, we score and rank answers by quantifying their relative positions in the semantic space.

    To achieve that, we regard the generated answers as nodes and the similarity between these answers as the weights of edges to construct a graph. In such a graph, we use the TextRank algorithm~\citep{mihalcea-tarau-2004-textrank} to calculate the weights of the nodes to reflect their ranks.
    Specifically, when computing the weight of an edge connecting two nodes $n_i$ and $n_j$, we utilize a Sentence-BERT~\citep{reimers-gurevych-2019-sentence} to embed their corresponding answers $a_i$ and $a_j$ to vectors, calculating the cosine similarity between the two vectors, as shown in eq.~\eqref{eq:sbert}:
    \begin{equation}
        S(n_i, n_j)=\frac{\text{SBERT}(a_i) \cdot \text{SBERT}(a_j)}{||\text{SBERT}(a_i)|| \  ||\text{SBERT}(a_j)||},
        \label{eq:sbert}
    \end{equation}
    where $\text{SBERT}(\cdot)$ is the embedding function of the Sentence-BERT.
    Subsequently, we calculated the weight of a node $n_i$ as follows:
   \begin{equation}
        W(n_i)=\sum_{n_j \in \text{I}(n_j)}\frac{S(n_i, n_j)}{\sum_{n_k \in \text{O}(n_j)}S(n_j, n_k)}W(n_j),
    \end{equation}
    where $\text{I}(n_i)$ represents the set of nodes that point to node $n_i$, and $\text{O}(n_j)$ represents the set of nodes that node $n_j$ points to.
    
     Following previous study~\citep{page1998pagerank}, we incorporate an empirical damping factor $d=0.85$ into the formulation to ensure the stability and convergence of the algorithm, that is:
       \begin{equation}
        W'(n_i)=(1-d)+d*W(n_i).
        \label{eq:weights}
    \end{equation}
    
    The overall ranking algorithm is shown as Algorithm~\ref{al:textrank}. 
    \begin{algorithm}[h]
        \caption{TextRank Algorithm}
        \label{al:textrank}
        \begin{algorithmic}[1]
            \REQUIRE ~~\\ 
                A set of $m$ nodes $(n_1,... ,n_m)$ corresponding to answers $(a_1,... ,a_m)$;\\
            \ENSURE ~~\\ 
                Ranking of answers $[a_1,... ,a_m]$;
            \STATE Initialize all the node weights $W(n)$ to 1.0;\
            \FOR{each $i \in [1, m-1]$}
                \FOR{each $j \in [i+1, m]$}
                    \STATE Calculate the similarity between $n_i$ and $n_j$ according to Eq.~\eqref{eq:sbert};
                \ENDFOR
            \ENDFOR
            \FOR{each $i \in [1, m]$}
                \STATE Calculate the weight of node $n_i$ using Eq.~\eqref{eq:weights};
            \ENDFOR
            \STATE Rank the nodes in descending order according to their weights.
        \end{algorithmic}
    \end{algorithm}

\subsection{How to construct contrasting data?}
     After we rank all the answers, we only extract a subset from them to train a reward model. Our motivation is that the PLM may generate duplicate or similar outputs. For instance:
    
    \textbf{Question:} \textit{How to improve concentration?}
    
    \textbf{Answer A:} \textit{Minimize distractions, use focus techniques, and manage time effectively.}
    
    \textbf{Answer B:} \textit{Avoid interruptions, apply concentration methods, and utilize time management skills.}

    In this example, learning the relative ranking of these two answers is irrational as they convey the same meaning. Therefore, we first cluster answers by minimizing the semantic distance within the clusters. Then, we retain only one representative answer within each cluster.

    However, it is difficult to pre-determine the optimal number of clusters in real-world cases. For instance, in multiple-choice questions, the optimal number of clusters is expected to be similar to the number of options. 
    On the contrary, open-ended questions may require more clusters to capture the diversity of answers.
    To solve this problem, we used the ISODATA to select the representative subset. The advantage of employing ISODATA is its ability to adaptively adjust the number of answer clusters by merging or splitting them for different questions, as shown in Algorithm~\ref{al:isodata}. Please note that the effectiveness of TextRank relies on a substantial number of samples. Therefore, we should only filter answers after ranking them.
    \begin{algorithm}[h]
        \caption{ISODATA Algorithm}
        \label{al:isodata}
        \begin{algorithmic}[1]
            \REQUIRE ~~\\ 
                A set of answers $A=(a_1,... ,a_m)$;\\
            \ENSURE ~~\\ 
                A subset of $A$: $[a'_1,... ,a'_{m'}]$;
            \STATE Randomly initialize $K$ cluster $C=(c_1,... ,c_k)$.
            \WHILE{Convergence criteria are not reached}
                \FOR{each $i \in [1, m]$}
                    \STATE Assign $a_i$ to the nearest cluster.\
                \ENDFOR
                \STATE Calculate the centroid of each cluster.
                \WHILE{the number of samples in the cluster $c_i$ $\le$ min-threshold, $c_i \in C$}
                        \STATE Merge $c_i$ and $c_j$, where $c_j$ is the closest cluster for $c_i$.
                \ENDWHILE
                \WHILE{the number of samples in the cluster $c_i$ $\ge$ max-threshold, $c_i \in C$}
                        \STATE Calculate the distance between each sample point in $c_i$ and the cluster centroid, split the farthest sample point.
                \ENDWHILE
            \ENDWHILE
        \STATE Extract centroids of all clusters as outputs. 
        \end{algorithmic}
    \end{algorithm}
    
    After the clustering, we build high-low-ranked answer pairs to train a reward model. To increase the quality difference between the two answers in an answer pair, we propose to select them with a fixed interval. Let $A'=(a'1,...,a'{m'})$ be a ranked answer set from ISODATA, the $i$'th answer pair would be $\text{AP}_i=(a'w, a'{l})$, where $l \geq w+\text{IL}$, and IL is the interval length, i.e. distances in the ranking list.

    Furthermore, we propose noise injection to the low-ranked answers to ensure the correctness of the ranking. Here, we designed three types of noise injection: 
    1) n-gram level editing operations involve randomly deleting or replacing an n-gram (where n < 4) with another random n-gram or inserting a random n-gram to the answer. 
    2) Adding or deleting negation words, where we randomly add negation words before verbs and delete them for answers that already contain negation words. 
    3) We consider randomly shuffling the order of sentences for answers that contain more than one sentence. When a low-ranked answer is identified for noise injection, we randomly select one of the three noise types. 
   Here, the noise injection is employed to enhance the diversity in the quality of the two answers within an answer pair. Moreover, counterexamples generated based on the above three manually defined rules can assist in training the language model to mitigate the severity of corresponding errors, thereby improving the overall generation quality, as demonstrated in the experimental section below.

\subsection{Training}
   After constructing answer pairs, we fine-tune another PLM as the reward model to convert answer ranks into numerical values to represent their reasonableness. 
    In detail, during the training process of the reward model, we maximize the reward difference between the two answers $(a_w, a_l)$ in an answer pair for a question $q$:
    \begin{equation}
        R_\theta(a_w, a_l; q)=\sigma(r_\theta(q, a_w) - r_\theta(q, a_l)).
    \end{equation}
    The loss function of the reward model $\theta$ is:
    \begin{equation}
        \text{loss}(\theta)=-E_{(q,a_w,a_l) \in D}[\log R_\theta(a_w, a_l; q)],
    \end{equation}
    where $D$ is the set of answer pairs. 
    
    Regarding the training of our generative policy, we maximizing our objective function, as shown in eq.~\eqref{eq:ppo}. For a given input $x$ and a model output $y$, the objective function comprises two components: 
    The score computed by the reward model and the KL divergence between the generative policy $\pi_\phi^\text{RL}$ and a sampling policy $\pi^\text{SFT}$. 
    \begin{equation}
        \begin{aligned}
        \text{objective}(\phi)=E_{(x,y) \in D_{\pi_\phi^\text{RL}}}[R_\theta(x,y)\\
        -\beta \log(\pi_\phi^\text{RL}(y|x)/ \pi^\text{SFT}(y|x))],
        \end{aligned}
        \label{eq:ppo}
    \end{equation}
    where the sampling policy $\pi^\text{SFT}$ is an original copy of the generative policy.

    Through this approach, our generative policy generates answers with high rewards while preserving its original question-answering capabilities~~\citep{ouyang2022training}.

\section{Experiments}
\label{sec:exp}
    To validate the effectiveness of our approach, we conducted experiments on three tasks: dialogue, story generation, and natural language understanding (NLU). To enhance the uniformity of our evaluation, we consider them all as question-answering tasks with different prompts~\citep{qi-etal-2022-enhancing}. Furthermore, we employed three human annotators to rank model outputs on two datasets with open-domain answers, computing the similarity between the results of our annotations and human annotations.

\subsection{Data sets used}
    \paragraph{DailyDialogue} is an extensive English conversation dataset that covers various topics and is collected from English learning websites~\citep{li-etal-2017-dailydialog}. 
    The conversations are authored by English speakers and showcase a natural language style with high complexity and diversity.
    
    \paragraph{The Cornell Movie-Dialogue Corpus} is an English movie dataset collected by Cornell University~\citep{Danescu-Niculescu-Mizil+Lee:11a}. 
    The dataset includes dialogues between movie characters, covering the conversations from more than 600 movies. These dialogues cover romantic, sci-fi, and thriller films. 
    
    \paragraph{Stanford Question-Answering Dataset v2.0 (SQuAD)} is a reading comprehension dataset, which comprises a question set raised by crowdworkers on a variety of Wikipedia articles~\citep{rajpurkar-etal-2016-squad}. For each question, the answer is a text segment, also known as a span, obtained from the corresponding reading passage.

    \begin{table*}[]
        \centering
        \resizebox{\textwidth}{!}{
        \begin{tabular}{llrrrlrrr}
        \thickhline
        Dataset      &  & \multicolumn{3}{c}{DailyDialogue}                                                        &  & \multicolumn{3}{c}{CornellMovie}                                                        \\ \cline{3-5} \cline{7-9} 
        Models      &  & \multicolumn{1}{c}{BLEU $ \uparrow$} & \multicolumn{1}{c}{GLEU $ \uparrow$} & \multicolumn{1}{c}{METEOR $ \uparrow$} &  & \multicolumn{1}{c}{BLEU  $ \uparrow$} & \multicolumn{1}{c}{GLEU $ \uparrow$} & \multicolumn{1}{c}{METEOR $ \uparrow$} \\ \hline
        Original  &  & 1.00 $\pm$ 0.15       & 2.33 $\pm$ 0.23        & 6.53 $\pm$ 0.52       &  & 3.25 $\pm$ 0.30        & 6.92 $\pm$ 0.27      & 15.76  $\pm$ 0.38        \\
        Full fine-tuning  &  & 1.99 $\pm$ 0.22       & 4.42 $\pm$ 0.39         & 9.82 $\pm$ 0.63    &  & 7.75 $\pm$ 0.37      & 12.72 $\pm$ 0.34        & 20.97 $\pm$ 0.48                     \\
        Adapter~\citep{lin-etal-2020-exploring} &  & 1.79 $\pm$ 0.18       & 4.13 $\pm$ 0.28         & 9.79 $\pm$ 0.74    &  & 7.74 $\pm$ 0.36      & 12.62 $\pm$ 0.42        & 20.92 $\pm$ 0.56    \\
        BitFit~\citep{ben-zaken-etal-2022-bitfit} &  & 1.87 $\pm$ 0.23       & 3.96 $\pm$ 0.43         & 9.92 $\pm$ 0.93    &  & 7.88 $\pm$ 0.37      & 13.02 $\pm$ 0.94        & 22.76 $\pm$ 0.53                    \\
        LoRA~\citep{hu2022lora} &  & 2.10 $\pm$ 0.41       & 4.89 $\pm$ 0.54         & \textbf{10.98 $\pm$ 1.03}    &  & 6.93 $\pm$ 0.28      & 12.02 $\pm$ 0.94        & 20.76 $\pm$ 0.33                     \\
        \rowcolor{lightgray} Ours     &  & 2.12 $\pm$ 0.33                 & 4.72 $\pm$ 0.68                 & 10.52 $\pm$ 1.08                   &  & 8.93 $\pm$ 1.05                 & 12.09 $\pm$ 1.01   & 21.58 $\pm$ 1.31                                \\ 
        \rowcolor{lightgray} Ours (with adding Noise)  &  & \textbf{2.40 $\pm$ 0.28}       & \textbf{4.90 $\pm$ 0.41}        & 10.77 $\pm$ 0.68       &  & \textbf{10.12 $\pm$ 1.02}        & \textbf{14.08 $\pm$ 0.96}      & \textbf{27.05  $\pm$ 1.12}        \\\thickhline
        \end{tabular}
        }
        \caption{The evaluation results with GPT-2 on two data sets (open-domain tasks) with 0.95 confidence level. `Bitfit' stands for Bias-only fitting. `LoRA' stands for Low-rank adaptation.}
        \label{tab:opendomain2}
    \end{table*}

        \begin{table*}[]
        \centering
        \resizebox{\textwidth}{!}{
        \begin{tabular}{llrrrlrrr}
        \thickhline
        Dataset      &  & \multicolumn{3}{c}{DailyDialogue}                                                        &  & \multicolumn{3}{c}{CornellMovie}                                                        \\ \cline{3-5} \cline{7-9} 
        Models      &  & \multicolumn{1}{c}{BLEU $ \uparrow$} & \multicolumn{1}{c}{GLEU $ \uparrow$} & \multicolumn{1}{c}{METEOR $ \uparrow$} &  & \multicolumn{1}{c}{BLEU  $ \uparrow$} & \multicolumn{1}{c}{GLEU $ \uparrow$} & \multicolumn{1}{c}{METEOR $ \uparrow$} \\ \hline
        Original  &  & 1.21 $\pm$ 0.17       & 3.53 $\pm$ 0.40        & 7.14 $\pm$ 0.89       &  & 3.75 $\pm$ 0.28        & 7.75 $\pm$ 0.56      & 8.53  $\pm$ 0.46        \\
        Full fine-tuning  &  & 2.03 $\pm$ 0.30       & 5.72 $\pm$ 0.48         & 10.86 $\pm$ 0.90    &  & 8.38 $\pm$ 0.46      & 13.95 $\pm$ 1.04        & \textbf{25.68 $\pm$ 0.84}                     \\
        Adapter~\citep{lin-etal-2020-exploring} &  & 1.87 $\pm$ 0.28       & 5.33 $\pm$ 0.46         & 9.89 $\pm$ 0.66    &  & 7.74 $\pm$ 0.40      & 13.26 $\pm$ 0.50        & 21.87 $\pm$ 0.74    \\
        BitFit~\citep{ben-zaken-etal-2022-bitfit} &  & 2.12 $\pm$ 0.32       & 5.85 $\pm$ 0.62         & \textbf{11.92 $\pm$ 1.21}    &  & 8.46 $\pm$ 0.46      & 14.12 $\pm$ 1.25        & 23.64 $\pm$ 0.72                    \\
        LoRA~\citep{hu2022lora} &  & \textbf{2.52 $\pm$ 0.83}       & 6.30 $\pm$ 0.54         & 11.89 $\pm$ 1.45    &  & 8.27 $\pm$ 0.36      & 13.58 $\pm$ 1.17        & 23.52 $\pm$ 0.45                     \\
        \rowcolor{lightgray} Ours    &  & 2.26 $\pm$ 0.75                 & 6.28 $\pm$ 0.96                 & 11.27 $\pm$ 1.87                   &  & 10.46 $\pm$ 2.01                 & 14.13 $\pm$ 2.53   & 23.28 $\pm$ 3.07                                \\ 
        \rowcolor{lightgray} Ours (with adding Noise)  &  & 2.33 $\pm$ 0.67       & \textbf{6.86 $\pm$ 0.76}        & 11.78 $\pm$ 0.84       &  & \textbf{10.96 $\pm$ 2.00}        & \textbf{14.64 $\pm$ 2.17}      & 23.83  $\pm$ 2.55        \\\thickhline
        \end{tabular}
        }
        \caption{The evaluation results with GPT-Neo on two data sets with 0.95 confidence level. }
        \label{tab:opendomainneo}
    \end{table*}
    
\subsection{Details}
    In this paper, we carried out experiments with the GPT-2 with 124 million parameters and GPT-Neo with 125 million parameters. \footnote{\href{https://github.com/ShuoYangtum/STA/tree/main}{Our code is available at GitHub.}}
    
    In terms of the TextRank, we set the maximum iterations to 1,000. For ISODATA, we set the cluster splitting variance threshold to 0.05.
    As for training, we used a mini-batch size of 16, and the optimizer is AdamW \citep{loshchilov2018decoupled} with a learning rate of $3e-5$. In addition, we set the value of $\beta$ to 0.5 in eq.~\eqref{eq:ppo}. Regarding inference, we controlled the maximum generated length below 100 tokens; the top-p value is 0.95 and the temperature is 0.8.

    To ensure the model's generalization, we conducted all dataset fine-tuning in an autoregressive method of standard language models. However, this fine-tuning approach may lead to lower performance in our experiments than those reported in works focused on specific domains.
    
\begin{table*}[h]
\centering
\resizebox{\textwidth}{!}{
\begin{tabular}{lrrrrrrr}
    \thickhline
    Dataset                    & \multicolumn{6}{c}{SQuAD v2.0}                                                      \\ \cline{2-8} 
    Models                           & \multicolumn{1}{c}{BLEU $ \uparrow$} & \multicolumn{1}{c}{GLEU  $ \uparrow$}       & \multicolumn{1}{c}{METEOR  $ \uparrow$}     & \multicolumn{1}{c}{EM  $ \uparrow$} & \multicolumn{1}{c}{Precision  $ \uparrow$} & \multicolumn{1}{c}{Recall  $ \uparrow$} & \multicolumn{1}{c}{F1  $ \uparrow$} \\ \hline
    Original                     & 2.70$\pm$0.75               & 4.07$\pm$1.01  & 11.66$\pm$2.17 & 10.40\%  & 0.06\%  & 0.12\%   & 0.08                   \\
    Full fine-tuning           & 19.56$\pm$2.67              & 39.75$\pm$4.23 & 38.24$\pm$3.75 & 52.60\%    & 50.64\%      & 51.94\%  & 0.51               \\
    Adapter~\citep{lin-etal-2020-exploring}  & 18.32$\pm$1.63              & 38.77$\pm$4.84 & 38.20$\pm$3.26 & 52.40\%      & 50.76\%           & 51.38\%  & 0.51       \\
    BitFit~\citep{ben-zaken-etal-2022-bitfit}  & 18.42$\pm$1.99              & 36.27$\pm$3.58 & 38.40$\pm$3.77 & 53.20\%      & 50.82\%           & 51.63\%  & 0.51       \\
    LoRA~\citep{hu2022lora} & \textbf{19.67$\pm$1.68}              & 40.42$\pm$3.98 & 38.24$\pm$3.75 & 54.40\%        & 52.63\%     & 53.12\%     & 0.53        \\
    \rowcolor{lightgray} Ours & 19.33$\pm$2.64              & 40.00$\pm$4.25 & 38.44$\pm$3.76 & 55.60\%      & 53.87\%       & 52.19\%   & 0.53           \\
    \rowcolor{lightgray} Ours (with adding Noise)    & 19.47$\pm$2.61              & \textbf{41.81$\pm$4.28} & \textbf{38.45$\pm$3.75} & \textbf{56.80\%}    & \textbf{54.23\%}  & \textbf{54.61\%}   & \textbf{0.54}                \\ \thickhline
\end{tabular}
}
\caption{The test results with GPT-2 on SQuAD v2.0 (NLU task) with 0.95 confidence level.}
\label{tab:squad2}
\end{table*}

\begin{table*}[h]
\centering
\resizebox{\textwidth}{!}{
\begin{tabular}{lrrrrrrr}
    \thickhline
    Dataset                    & \multicolumn{6}{c}{SQuAD v2.0}                                                      \\ \cline{2-8} 
    Models                           & \multicolumn{1}{c}{BLEU $ \uparrow$} & \multicolumn{1}{c}{GLEU  $ \uparrow$}       & \multicolumn{1}{c}{METEOR  $ \uparrow$}     & \multicolumn{1}{c}{EM  $ \uparrow$} & \multicolumn{1}{c}{Precision  $ \uparrow$} & \multicolumn{1}{c}{Recall  $ \uparrow$} & \multicolumn{1}{c}{F1  $ \uparrow$} \\ \hline
    Original                     & 3.95$\pm$0.88               & 7.64$\pm$2.12  & 23.23$\pm$3.98 & 12.20\%  & 0.10\%  & 0.18\%   & 0.12                   \\
    Full fine-tuning           & 22.23$\pm$4.00              & 41.76$\pm$5.68 & 40.72$\pm$4.25 & 54.20\%    & 53.82\%      & 52.97\%  & 0.53               \\
    Adapter~\citep{lin-etal-2020-exploring}  & 16.97$\pm$2.33              & 34.98$\pm$5.60 & 38.05$\pm$4.62 & 52.40\%      & 51.66\%           & 52.88\%  & 0.52       \\
    BitFit~\citep{ben-zaken-etal-2022-bitfit}  & 18.42$\pm$2.00              & 40.46$\pm$2.87 & 39.96$\pm$3.53 & 53.40\%      & 53.20\%           & 50.65\%  & 0.52       \\
    LoRA~\citep{hu2022lora} & 22.58$\pm$3.66              & \textbf{43.82$\pm$4.38} & 40.24$\pm$4.75 & 55.80\%        & \textbf{54.80\%}     & 55.20\%     & \textbf{0.55}        \\
    \rowcolor{lightgray} Ours & 22.16$\pm$3.46              & 42.05$\pm$4.88 & \textbf{41.04$\pm$3.66} & 56.00\%      & 53.67\%       & 54.28\%   & 0.54           \\
    \rowcolor{lightgray} Ours (with adding Noise)    & \textbf{22.67$\pm$3.73}              & 41.88$\pm$5.32 & \textbf{41.04$\pm$3.46} & \textbf{57.20\%}    & 54.76\%  & \textbf{55.25\%}   & \textbf{0.55}                \\ \thickhline
\end{tabular}
}
\caption{The test results with GPT-Neo on SQuAD v2.0 (NLU task) with 0.95 confidence level.}
\label{tab:squadneo}
\end{table*}

\begin{comment}
\subsection{Prompt used}
In our experiments, we transformed different tasks into question-answer pairs using prompts, as shown in table~\ref{tab:prompt}.

\begin{table}[H]
\resizebox{\columnwidth}{!}{
\begin{tabular}{ll}
\hline
Dataset       & \multicolumn{1}{c}{Prompt}                                             \\ \thickhline
DailyDialogue & \begin{tabular}[c]{@{}l@{}}The following a chat log, please predict \\ the next response based on its context. \\ Chat log: \textless{}Q\textgreater{}. The next response is:\end{tabular}                             \\ \hline
CornellMovie  & \begin{tabular}[c]{@{}l@{}}The following is a movie dialogue, \\ please continue the script \\ with 50-100 words based on the dialogue. \\ Dialogue: \textless{}Q\textgreater{}. The following script is:\end{tabular} \\ \hline
SQuAD v2.0    & \begin{tabular}[c]{@{}l@{}}Please read the following text \\ to answer the question at the \\ end of the passage: \textless{}Q\textgreater{}. The answer is:\end{tabular}                                              \\ \thickhline
\end{tabular}
}
\caption{The dataset used in the experiment and the corresponding prompt. <Q> represents the input text.}
\label{tab:prompt}
\end{table}
\end{comment}

\subsection{Automated evaluation eMetric}
\label{sec:evaluation}
    \paragraph{Exact Match Score (EM)} is used to evaluate the prediction accuracy of a classification model. 
    It refers to the proportion of questions for which the model provides the correct answer.
    
    \paragraph{The Bilingual Evaluation
    Understud (BLEU)} is a metric used to evaluate the quality of generation~\citep{papineni-etal-2002-bleu}. 
    We used the BLEU-4 metric via NLTK~\citep{bird2009natural} to quantitatively assess the similarity between machine outputs and human reference.
    
    \paragraph{GLEU} is designed to estimate text fluency solely based on parser outputs~\citep{mutton-etal-2007-gleu}. 
    The metric was examined by analyzing its correlation with human judgments of text fluency. 
    
    \paragraph{METEOR} is a metric based on word-level exact matching~\citep{banerjee-lavie-2005-meteor}. It considers factors such as lexical overlap, word order differences, and stem changes between the generated text and the reference text. 

\subsection{Manual evaluation}
    We employed three human annotators through the crowdsourcing platform of MolarData to validate our hypothesis. We randomly selected 100 questions from each of the two open-ended question-answering datasets used and constructed four answers for each question. Our evaluation includes 1. the similarity between human and automated ranking and 2. the kappa~\citep{cohen1960coefficient} consistency coefficient among different annotators.

\section{Results and Discussion}
\label{sec:result}

\subsection{Analysis}
\label{sec:analysis}
    \begin{comment}
Table~\ref{tab:reward} presents the strong correlation between answer rank and quality in our pipeline-generated comparative data. The ranking accuracy was evident in all three datasets, exhibiting clear stepwise patterns, demonstrating the effectiveness of our proposal. Notably, the ranking accuracy was most apparent in the SQuAD dataset, which has shorter answers. This feature makes it more likely for higher-ranked answers to match the ground truth, resulting in higher-ranking consistency.

Furthermore, we observed that the METEOR tended to be higher than the BLEU, indicating that our generated answers have a greater probability of semantic alignment with the reference answers, even if they do not match the reference answers at the token level, as METEOR considers sentence-level semantic similarity.
    \end{comment}
    
Table~\ref{tab:opendomain2} and Table~\ref{tab:opendomainneo} present the comparative results of our fine-tuning method in the open-domain question-answering task against the baselines. Our two models achieved state-of-the-art (SOTA) results in almost all evaluation metrics. Regarding the GPT-2 model, our method performed similarly to the LoRA algorithm while outperforming the direct fine-tuning and BitFit algorithm by nearly 1 point in three evaluation metrics on the Dialogue dataset. Additionally, on the CornellMovie dataset, our method achieved a significant improvement of almost 30\% in BLEU and METEOR. Furthermore, we surpassed the BitFit algorithm by approximately five points in the text fluency metric GLEU.
Moreover, we discovered that introducing noise during the training of the reward model led to improved performance. We attribute this enhancement to the additional contrastive information brought by the noise, which prevented the model from generating incorrect answers to some extent. For the GPT-Neo model, our model outperformed the baselines with a relatively modest advantage. This may be attributed to GPT-Neo having more parameters and better performance than GPT-2, resulting in a lower probability of generating incorrect answers and thus benefiting less from our self-correction approach. Furthermore, the GPT-Neo demonstrated superior performance to the GPT-2, which may be attributed to its larger size and pre-training data scale.

Table~\ref{tab:squad2} and Table~\ref{tab:squadneo} present the measurement results on the SQuAD dataset. Our models outperformed baselines by approximately two points in the EM metric and slightly surpassed the LoRA algorithm in the F1 metric. Additionally, our model outperformed baselines regarding text fluency and semantic similarity with standard answers. Furthermore, we found that our models exhibited significant improvement in the SQuAD dataset. We suppose this is because the SQuAD dataset has deterministic answers, making it easier for the model to learn a deterministic mapping function and achieve better performance during fine-tuning. These findings indicate that our method applies not only to dialogue systems but also brings improvements in general NLP tasks.

        \begin{figure}[ht]
        \centering
            \includegraphics[width=7cm]{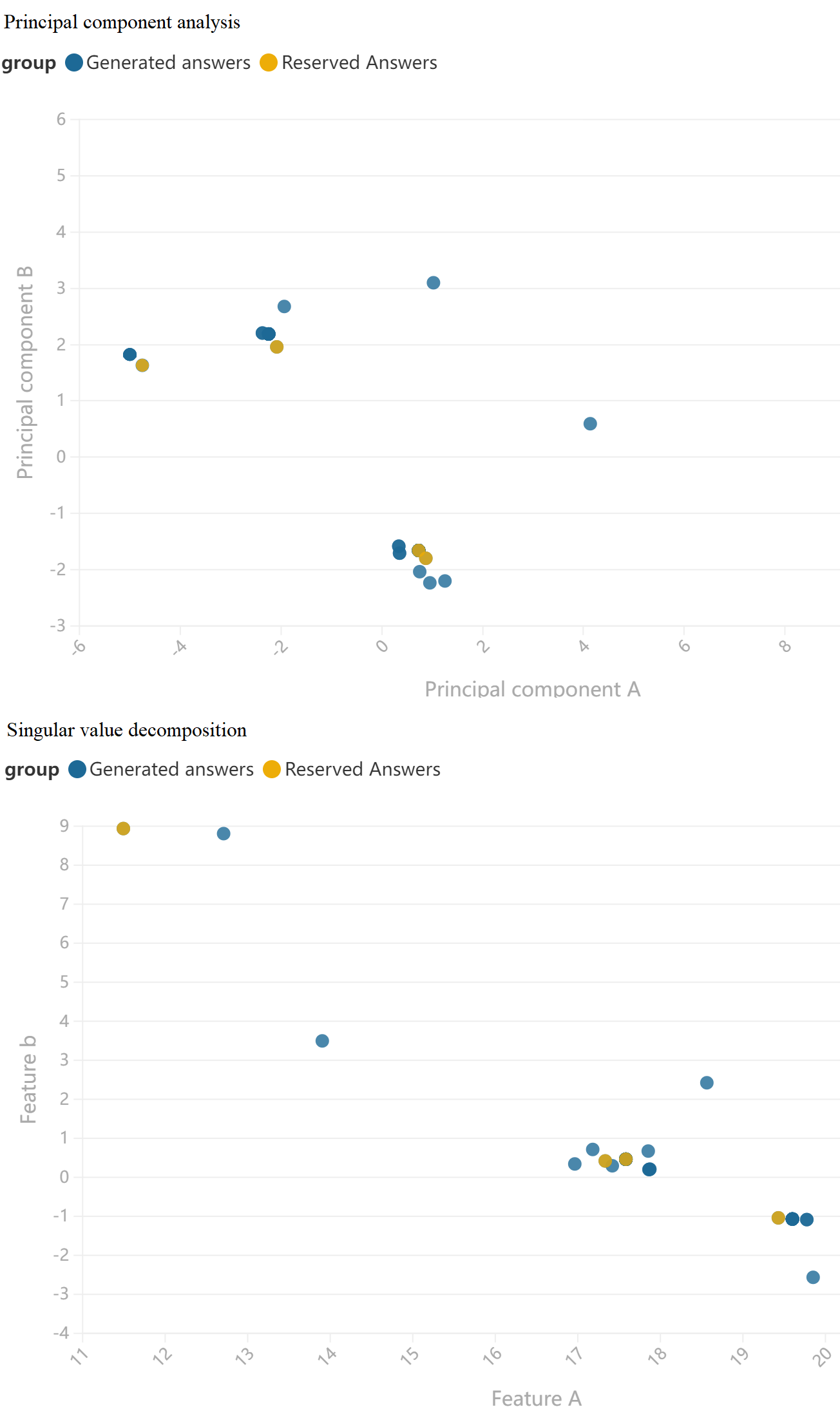}
            \caption{We applied principal component analysis and singular value decomposition to the BERT embeddings of answers generated by GPT-2.}
            \label{fig:pca}
    \end{figure}
    
 \subsection{Analysis of manual evaluation}
 Due to the relatively poor performance of our GPT-2-based generative policy, we conducted a manual evaluation for the ability of answer ranking of the reward model used. Our manual evaluation demonstrated that the reward model achieved an 83.33\% probability of ranking answers in the same order as the human beings on the DailyDialogue dataset, and a 63.00\% of that was observed on the Movie dataset. Notably, the figure of that was only 4.17\% for random ranking. Consequently, our reward model exhibited a high level of consistency with human judgments of answer quality and can serve as a substitute for manual annotation to a significant extent.

 Furthermore, as an ablation study, we observed that human annotators achieved higher agreement coefficients when ranking answers generated with ISODATA. Specifically, on the DailyDialog dataset, the kappa coefficient of the three annotators for ranking filtered answers was 46.67. In contrast, the kappa coefficient for unfiltered answers was only 26.67. Similarly, on the CornellMovie dataset, the corresponding kappa coefficients for filtered and unfiltered answers were 18.51 and 11.11, respectively. Our analysis suggests that using the ISODATA algorithm can enhance the representativeness of the answers generated by our method.
\subsection{Additional study}
    \paragraph{Can we observe clusters of answers in the semantic space?}
    We conducted an additional experiment with GPT-2 to validate our hypothesis in Section~\ref{sec:rank}, i.e., clusters of answers generated exist in the semantic space. Specifically, we randomly selected a question from the SQuAD dataset and applied principal component analysis~\citep{pearson1901liii} and singular value decomposition~\citep{golub1965calculating} to visualize semantic vectors of answers generated, as illustrated in Figure~\ref{fig:pca}. Here, we used a BERT model~\citep{Devlin2019BERTPO} not utilized in the above experiments to embed these answers to prevent possible information leakage.

    Our analysis indicates that most of the answers generated exhibit apparent clustering. At the same time, the scattered data points are irrelevant or incorrect answers generated with relatively high probabilities during decoding. These low-quality answers were ranked lower as negative samples for training the reward model. Furthermore, we observed that the cluster centers obtained by the ISODATA algorithm can represent the distribution pattern of the answers.

\section{Limitations}
    While our model outperforms the baseline on various evaluation metrics, our fine-tuning approach first exhibits higher computational complexity. Regarding space complexity, the proposed method requires using three pre-trained models simultaneously when applying the PPO: a reward model, a PLM for the generative policy, and another PLM for the sampling policy. Since the space complexity counts three times that of the fine-tuned PLMs, our method is much more computationally demanding than the baselines.
    Regarding the time complexity, all fine-tuning approaches presented in this study exhibit a linear one, i.e., proportional to the sequence's length. However, in practical training, our training time is more than twice that required for full fine-tuning since the input sequences should be input to three models for loss computation. Compared to non-full-parameter fine-tuning methods like LoRA, although our approach has improved performance, it may also require more noticeable resource consumption.
    
    Secondly, the extensive knowledge base of large language models may result in not generating answers with noticeable deviations, i.e., error points, in the semantic distribution when developing answers. This also makes the three noise introduction rules we proposed unable to create negative samples that could significantly improve their performance. Due to limitations in our computational resources, we can only make theoretical assumptions the proposed approach yields performance improvements when we apply it to advanced large-scale PLMs. In summary, while the limitations exist, our approach addressed the reliance on human labor in the RLHF training pipeline and demonstrated its effectiveness on popular PLMs with relatively fewer parameters.
    
\section{Conclusion}
    This paper introduces STR, a self-supervised pipeline that leverages proximal policy optimization for fine-tuning language models. We aim to reduce the need for manual labor, making RLHF-based algorithms more accessible and practical for researchers. Experimental results with two models across three tasks show that our fine-tuning method improves three points over the baselines regarding BLEU, ROUGE, and METEOR. Additionally, manual evaluations indicate that our proposed text ranking algorithm generates annotations similar to human-generated ones. This discovery provides a cost-effective framework for future PPO-guided models to automatically generate training data. As a result, our research contributes to the advancement of self-supervised learning in fine-tuning pre-trained language models, opening up new possibilities for applying reinforcement learning in natural language processing.

\begin{comment}
\section{Ethics Statement}
    As part of our submission, we present an Ethics Statement regarding our research and experimental practices. 

    Firstly, all the datasets and code libraries used in our experiments are open-sourced and have been widely used in previous academic literature published. Secondly, we employed human annotators for manual evaluations through legitimate crowdsourcing platforms. These annotators comprehend our task and the purpose behind our data collection. They accepted the compensation agreed upon by both parties beforehand. Lastly, all individuals involved in the evaluation were kept anonymous to our authors.    
    
    The above practices have ensured the ethical conduct of our research and the integrity of our findings. We are committed to upholding the highest standards of research ethics and transparency in our research.
    
\end{comment}

\bibliography{anthology,custom}
\bibliographystyle{acl_natbib}

\appendix

\end{document}